\documentclass[12pt]{article}

\usepackage[bitstream-charter]{mathdesign}
\usepackage[mathcal]{eucal}

\usepackage{float}
\usepackage{algorithm}
\usepackage{amsmath}
\usepackage{graphicx}
\usepackage{graphicx}
\usepackage{array}
\usepackage{tabularx}
\usepackage{color,epsfig}
\usepackage{cite}
\usepackage{url}
\usepackage[colorlinks=true,citecolor=blue]{hyperref}
\usepackage{a4wide}
\usepackage[framemethod=tikz]{mdframed}
\usepackage{booktabs}
\usepackage{multirow}
\usepackage{color}
\usepackage{colortbl}
\usepackage{subcaption}
\usepackage{fancyhdr}

\fancypagestyle{titlepage}{%
    \fancyhf{}
  \rhead{Published on \textbf{CAAI Transactions on Intelligence Technology}}
}

\definecolor{mycolor}{rgb}{0.122, 0.435, 0.698}

\newcommand{\ra}[1]{\renewcommand{\arraystretch}{#1}}

\begin{document}

\title{Compressing deep quaternion neural networks with targeted regularization}

\author{Riccardo~Vecchi, Simone Scardapane, \\ Danilo Comminiello, and Simone Scardapane\thanks{Authors are with the Department of Information Engineering, Electronics, and Telecommunications, Sapienza University of Rome, Via Eudossiana 18, 00184, Rome, Italy.
E-mail: \{firstname.lastname\}@uniroma1.it.}}

\maketitle
\thispagestyle{titlepage}

\begin{abstract}
In recent years, hyper-complex deep networks (such as complex-valued and quaternion-valued neural networks) have received a renewed interest in the literature. They find applications in multiple fields, ranging from image reconstruction to 3D audio processing. Similar to their real-valued counterparts, quaternion neural networks (QVNNs) require custom regularization strategies to avoid overfitting. In addition, for many real-world applications and embedded implementations, there is the need of designing sufficiently compact networks, with few weights and neurons. However, the problem of regularizing and/or sparsifying QVNNs has not been properly addressed in the literature as of now. In this paper, we show how to address both problems by designing targeted regularization strategies, which are able to minimize the number of connections and neurons of the network during training. To this end, we investigate two extensions of $\ell_1$ and structured regularization to the quaternion domain. In our experimental evaluation, we show that these tailored strategies significantly outperform classical (real-valued) regularization approaches, resulting in small networks especially suitable for low-power and real-time applications.
\end{abstract}

\section{Introduction}

Deep neural networks have achieved remarkable results in a variety of tasks and applicative scenarios over the last years \cite{basavegowda2020deep}. Several breakthroughs have been obtained by designing custom neural modules for exploiting structure in data, such as the spatial organization of pixels in an image, temporal or sequential information, and so on. The field of quaternion deep learning aims at extending these results to problems for which a hyper-complex representation (as opposed to a real-valued representation) is more adequate \cite{zhu2018quaternion,parcollet2018quaternion} (see also \cite{arena1997multilayer} for earlier works on the field). Among these approaches, the most common is the use of quaternion values (a straightforward extension of the complex algebra) for representing input values, weights, and output values of the network. The resulting quaternion-valued neural networks (QVNNs) have been successfully applied to, among others, image classification \cite{zhu2018quaternion,parcollet2019quaternion,chen2019low}, image coloring and forensics \cite{yin2019quaternion}, natural language processing \cite{tay2019lightweight}, graph embeddings \cite{zhang2019quaternion}, human motion recognition \cite{pavllo2019modeling}, and 3D audio processing \cite{comminiello2019quaternion}. By exploiting the properties of the quaternion algebra, QVNNs can achieve similar or higher accuracy than their real-valued counterparts, while requiring fewer parameters and computations. In fact, outside of the deep learning field, the use of quaternion representations is well established in, e.g., the robotics community for linear filtering processing \cite{cheng2019mobile}.

\pagestyle{empty}

The majority of the literature on QVNNs up to this point has focused on extending standard deep learning operations, such as convolution \cite{zhu2018quaternion}, batch normalization \cite{yin2019quaternion}, or weight initialization \cite{parcollet2018quaternion}, to the quaternion domain. Less attention, however, has been devoted to properly extending other aspects of the training process, including accelerated optimization algorithms \cite{xu2015optimization} and regularization strategies. In particular, in many real-world scenarios (e.g., embedded devices) users need to take into careful consideration of complexity and computational costs, by making the networks as small as possible while maintaining a good degree of accuracy \cite{alippi2002selecting,koneru2019sparse}.

In the real-valued case, these constraints have been analysed in detail, and several strategies have been developed. Most commonly, compression of neural networks can be achieved while training by applying several regularization strategies, such as $\ell_2$, $\ell_1$, or group sparse norms \cite{wen2016learning,scardapane2017group,moini2017resource}, which can target either single weights or entire neurons. A direct extension of these strategies to the case of QVNN, as done in the current literature, applies them independently on the four components of each quaternion weight. However, in this paper, we argue and show experimentally later on that this trivial extension results in highly sub-optimal regularization procedures, which do not sufficiently exploit the properties of the quaternion algebra. In fact, the problem of sparsifying a quaternion is not restricted to neural networks, but it has received attention from other disciplines, most notably quaternion extensions of matching pursuit \cite{barthelemy2014sparse}, and compressive sensing \cite{badenska2017compressed}. To the best of our knowledge, however, almost no work has been devoted to the specific problem of targeting quaternion-valued sparsity in QVNNs, possibly hindering their application in certain applicative scenarios.

\subsection*{Contributions of the paper}

In this paper, we leverage on prior works on compression of real-valued networks and quaternion-valued norms to propose two \textit{targeted} regularization strategies for QVNNs, filling an important gap in the literature.

\begin{enumerate}
	\item The first regularization approach that we propose (Section \ref{subsec:l1_quaternion_regularization}) extends the use of $\ell_1$ regularization to consider a single quaternion weight as a unitary component, and it is akin to a structured form of regularization in the real-valued case. It allows to remove entire quaternion weights simultaneously, instead of each of its four components independently.
	\item The second strategy that we propose is instead defined at the level of a single quaternion neuron, extending ideas from \cite{gordon2018morphnet} to a quaternion domain, thus allowing to remove entire units from the network at once. We consider real-valued regularization to be applied on quaternion extensions of batch normalization (BN), in which every neuron is scaled by a single real-valued coefficient and is eventually removed by the optimization process. Thus, we achieve a better stability (thanks to the use of BN) and sparsity at the same time.
\end{enumerate} 

In our experimental evaluation, we show that these two proposed strategies significantly outperform the naive application of classical regularization strategies on two standard image recognition benchmarks. The resulting QVNNs are thus smaller (both in terms of neurons and weights) and require a markedly smaller computational footprint when run in inference mode, with up to 5x reductions in the number of connections and 3x speedups in the inference time.

\subsection*{Organization of the paper}
Section \ref{sec:preliminaries} recalls quaternion-algebra and QVNNs. Section \ref{sec:target_regularization_for_qvnns} describes our two proposed regularization strategies. We provide an experimental evaluation in Section \ref{sec:experimental_results}, concluding in Section \ref{sec:conclusions}.

\subsection*{Notation}

In the rest of the paper, the use of a subscript $\left\{i,j,k\right\}$ always refers to the respective imaginary component of the corresponding quaternion number, as explained in Section \ref{subsec:quaternion_algebra}. $x^*$ is the conjugate of $x$. We use bold letters, e.g., $\mathbf{x}$, for vectors, and uppercase bold letters, e.g., $\mathbf{X}$, for matrices. For indexing, we use a superscript $l$ to refer to the $l$th layer in a network, while we use bracket notation to refer to an element inside a set (e.g., $\mathbf{x}(n)$ is the $n$th input element of the training set).

\section{Preliminaries}
\label{sec:preliminaries}
\subsection{Quaternion algebra}
\label{subsec:quaternion_algebra}
A quaternion-valued number $x \in \mathbb{H}$ can be represented by a tuple of four real-valued numbers $(x_r, x_i, x_j, x_k) \in \mathbb{R}^4$ as \cite{ward2012quaternions}:
\begin{equation}
    x = x_r + ix_i + jx_j + kx_k \,,
\end{equation}

\noindent where the three imaginary units $i$, $j$, $k$ satisfy the fundamental axiom of quaternion algebra  $i^2 = j^2 = k^2 = ijk = -1$. Given two quaternions $x$ and $y$, we can define their sum as:
\begin{equation}
    z = (x_r + y_r) + i (x_i + y_i) + j(x_j + y_j) + k(x_k + y_k) \,,
\end{equation}

\noindent and similarly for multiplication by a real number. More importantly, the (Hamilton) product between the two quaternions is given by:

\begin{align}
    x \otimes y  = & \left( x_ry_r - x_iy_i - x_jy_j - x_ky_k \right) + \nonumber \\
    & i \left( x_ry_i + x_iy_r + x_j y_k + x_ky_j \right) + \nonumber \\
    & j \left( x_ry_j - x_iy_k + x_jy_r + x_ky_i \right) + \nonumber \\
    & k \left( x_ry_k + x_iy_j - x_jy_i + x_ky_r \right) \,.
    \label{eq:hamilton_product}
\end{align}

\noindent Note that the product is not commutative, setting apart quaternion algebra from its complex- and real-valued restrictions.

\subsection{Quaternion-valued neural networks}

QVNN are flexible models for transforming quaternion-valued vector inputs $\mathbf{x} \in \mathbb{H}^d$ to a desired target value $\mathbf{y}$, which in the majority of cases are real-valued (e.g., a probability distribution over a certain number of classes). A standard, fully-connected layer of a QVNN is given by:

\begin{equation}
    g(\mathbf{h}^{l+1}) = \sigma\left( \mathbf{W} \otimes \mathbf{h}^l + \mathbf{b} \right) \,,
    \label{eq:qvnn_layer}
\end{equation}

\noindent where $\mathbf{h}^l$ is the input to the layer, $\mathbf{W}$ is a quaternion-valued matrix of adaptable coefficients with components $\left(\mathbf{W}_r, \mathbf{W}_i, \mathbf{W}_j, \mathbf{W}_k\right)$ (and similarly for $\mathbf{b}$), $\otimes$ performs matrix-vector multiplication according to the Hamilton product in \eqref{eq:hamilton_product}, and $\sigma(\cdot)$ is a proper element-wise non-linearity. Similarly to the complex-valued case \cite{scardapane2018complex}, choosing an activation function is more challenging than for real-valued NNs, and most works adopt a split-wise approach where a real-valued function $\sigma_r$ is applied component-wise:

\begin{equation}
    \sigma(s) = \sigma_r(s_r) + i \sigma_r(s_i) + j \sigma_r(s_j) + k \sigma_r(s_k) \,,
    \label{eq:quaternion_split_activation_function}
\end{equation}

\noindent where $s$ is a generic activation value. Customarily, the input to the first layer is set to $\mathbf{h}^1 = \mathbf{x}$, while the output of the last layer is the desired target $\mathbf{h}^L = \mathbf{y}$. If the target is real-valued, one can transform $\mathbf{h}^L$ to a real-valued vector by taking the absolute value element-wise, and eventually apply one or more real-valued layers afterwards. In addition, \eqref{eq:qvnn_layer} can be easily extended to consider convolutive layers \cite{zhu2018quaternion} and recurrent formulations \cite{parcollet2018quaternion}.

\subsection{Optimization of QVNNs}

Now consider a generic QVNN $f(\mathbf{x})$ obtained by composing an arbitrary number of layers in the form of \eqref{eq:qvnn_layer} or its extensions. We receive a dataset of $N$ examples $\left\{\mathbf{x}(n), \mathbf{y}(n)\right\}_{n=1}^N$, and we train the network by optimizing:

\begin{equation}
    J(\theta) = \frac{1}{N} \sum_{n=1}^N l\left( \mathbf{y}(n), f(\mathbf{x}(n)) \right) + \lambda \cdot r(\theta) \,,
    \label{eq:cost_function}
\end{equation}

\noindent where $\theta$ is the set of all (quaternion-valued) parameters of the network, $l$ is a loss function (e.g., mean-squared error, cross-entropy loss), and $r$ is a regularization function weighted by a scalar $\lambda \geq 0$. Because the loss function in \eqref{eq:cost_function} is non-analytic, one has to resort to the generalized QR-calculus to define proper gradients for optimization \cite{xu2015optimization}. Luckily, these gradients coincide with the partial derivatives of \eqref{eq:cost_function} with respect to all the real-valued components of the quaternions, apart from a scale factor. For this reason, it is possible to optimize \eqref{eq:cost_function} using standard tools from stochastic optimization popular in the deep learning literature, such as Adam or momentum-based optimizers.

While most components described up to now have received considerable attention in the literature, the design of a correct regularization term $r(\cdot)$ in \eqref{eq:cost_function} has been mostly ignored, and it is the focus of the next section.

\section{Targeted regularization for QVNNs}
\label{sec:target_regularization_for_qvnns}

In the real-valued case, a classical choice for the regularizer $r(\cdot)$ in \eqref{eq:cost_function} is the $\ell_2$ norm. Whenever sparsity is desired, it can be replaced with the $\ell_1$ norm, or a proper group version acting at a neuron level \cite{wen2016learning,scardapane2017group}. In most implementations of QVNNs, these regularizers are applied element-wise on the four components of each quaternion weight. For example, $\ell_1$ regularization in this form can be written as:
\begin{equation}
r(\theta) = \sum_{w \in \theta} \big( \lvert w_r \rvert + \lvert w_i \rvert +\lvert w_j \rvert + \lvert w_k \rvert \big) \,.
\end{equation}
We argue that, because of the decoupling across the four components, this operation results in far less regularization and sparsity than one could expect. This is inconvenient both from the generalization point of view, and from an implementation perspective, where smaller, more compact networks are desired. In this section, we present two \textit{targeted} regularization strategies, acting on each quaternion as a unitary component, resulting in a more principled form of regularization for QVNNs.

\subsection{$\ell_1$ regularization for quaternion weights}
\label{subsec:l1_quaternion_regularization}

Given any weight $w \in \mathbb{H}$ of the QVNN (i.e., a single element of $\theta$ from \eqref{eq:cost_function}), the first method we explore is to regularize its norm as:

\begin{equation}
    r(w) = \frac{1}{Q} \sqrt{w_r^2 + w_i^2 + w_j^2 + w_k^2} = \frac{1}{Q} \sqrt{w^* \otimes w} \,,
    \label{eq:reg_1}
\end{equation}

\noindent where $Q$ is the number of weights in the network (allowing to split the influence of \eqref{eq:reg_1} on each weight with respect to the loss function \eqref{eq:cost_function}), and $w^*$ is the conjugate of $w$, i.e., $w^* = w_r - iw_i - jw_j - kw_k$ (combining the definition of a conjugate with \eqref{eq:hamilton_product}, and removing all terms except the first line, shows the second equality in \eqref{eq:reg_1}). 

This method can be seen as the natural extension of $\ell_1$ norm minimization on a quaternionic signal \cite{badenska2017compressed}. It is also equivalent to a structured form of sparsity  \cite{scardapane2017group}, where we group all the components of the quaternion $w$ together. As a result, minimizing \eqref{eq:reg_1} will tend to bring the entire quaternion weight to $0$, instead of each component independently (similarly to how structured sparsity in a real-valued network brings all the incoming or outgoing weights of the network towards zero together \cite{scardapane2017group}).

\subsection{Sparse regularization with quaternion batch normalization}
\label{subsec:sparse_regularization_with_quaternion_batch_normalization}

The method described in Section \ref{subsec:l1_quaternion_regularization} is effective for removing single quaternion weights, but in many real-world scenarios we also require a principled way to remove entire neurons during the training process \cite{scardapane2017group}. To this end, we investigate a hyper-complex extension of the technique originally proposed in \cite{gordon2018morphnet}. The basic idea is to compose each layer in \eqref{eq:qvnn_layer} with a batch normalization (BN) layer \cite{ioffe2015batch}, and then perform sparse regularization on the parameters of the BN layer, indirectly removing the original neurons in the network. We briefly recall that BN, originally introduced in \cite{ioffe2015batch}, allows each neuron to adapt the mean and variance of its activation values, with the general effect of stabilizing training and possibly simplifying optimization.

For implementing the BN model in the quaternion domain, we build on \cite{yin2019quaternion} after we consider a single output in \eqref{eq:qvnn_layer}, i.e., the quaternion-valued output of a single neuron in the network. During training, we observe a mini-batch of $B$ inputs $\mathbf{x}(1), \ldots, \mathbf{x}(B)$ (a subset of the full dataset) and corresponding outputs of the neuron $h(1), \ldots, h(B)$ (we do not use an index for the neuron for notational simplicity). We can compute the mean and variance of the mini-batch as:

\begin{align}
    \hat{\mu} & = \frac{1}{B} \sum_{n=1}^B \Bigl[ h_r(n) + i h_i(n) + j h_j(n) + k h_k(n) \Bigr] \,, \\
    \hat{\sigma}^2 & = \frac{1}{B} \sum_{n=1}^B \left( h(n) - \hat{\mu} \right) \otimes \left( h(n) - \hat{\mu} \right)^* \,.
\end{align}

\noindent These values are computed dinamically during training, while they are set to a fixed (pre-computed) value during inference. The output of the BN layer is defined as \cite{yin2019quaternion}:

\begin{equation}
    h_{\text{BN}}(n) = \left( \frac{h(n) - \hat{\mu}}{\sqrt{\hat{\sigma}^2 + \varepsilon}} \right)\gamma + \beta \,,
    \label{eq:bn_layer}
\end{equation}

\noindent where $\varepsilon$ is a small value added to ensure stability, while $\gamma \in \mathbb{R}$ and $\beta \in \mathbb{H}$ are trainable parameters initialized at $1$ and $0$ respectively. Key for our proposal, the $\gamma$ parameter in \eqref{eq:bn_layer} is real-valued, allowing us to apply standard real-valued regularization. In particular, similarl to \cite{gordon2018morphnet}, we apply (real-valued) $\ell_1$ regularization on the $\gamma$s, since pushing a single $\gamma$ to zero effectively allows us to remove the entire neuron in the QVNN. Thus, denoting with $\Gamma$ the set of all $\gamma$ parameters in the network, we regularize them as:
\begin{equation}
r(\Gamma) = \frac{1}{\lvert \Gamma \rvert}\sum_{\gamma \in \Gamma} \lvert \gamma \rvert \,,
\end{equation}
where $\lvert \Gamma \rvert$ is the cardinality of the set $\Gamma$.
\subsection{Mixed regularization strategies}
The strategies described in the previous sections are not exclusive, and we can explore several \textit{mixed} strategies with different regularization weights. In our experimental section, we consider combining the two strategies, as long as one of the two strategies are combined with a classical $\ell_1$ regularization to be applied independently on each component.
\section{Experimental results}
\label{sec:experimental_results}
\subsection{Experimental setup}
We evaluate our proposal on two quaternion-valued image recognition benchmarks taken from \cite{zhu2018quaternion}. Firstly, we use the standard MNIST dataset by converting every image pixel to a quaternion  with $0$ imaginary components, i.e., we encode one grey-valued pixel $g$ as $x = g + i0 + j0 + k0$. Secondly, we consider the more challenging CIFAR-10 dataset by converting its RGB representation to the three imaginary components of a pure quaternion with $0$ real part, i.e., we encode a single pixel with RGB values $(r,g,b)$ to $x = 0 + ir + jg + kb$.

Similar to previous literature, for MNIST, we use a quaternion convolutional network with two convolutive layers having $16$ and $32$ filters respectively, interleaved by (quaternion-valued) max-pooling operations. After the second convolutive layer we apply a dropout operation for regularization and a final quaternion fully connected layer for obtaining the class probabilities. For CIFAR-10, we increase this to five convolutive layers having respectively $32$, $64$, $128$, $256$, and $512$ filters. In this case, we also apply droput every two convolutive layers. Overall, the MNIST network has $\approx 10k$ parameters, while the CIFAR-10 network has $\approx 500k$ parameters.

All networks use ReLU applied component-wise as in \eqref{eq:quaternion_split_activation_function}. After the last layer, we take the absolute values of each output to obtain a real-valued score (equivalent to the classical logit value in a real-valued network), and we apply a softmax activation function to convert these to probabilities. The networks are trained to minimize the average cost with a cross-entropy loss function using the Adam optimization algorithm.

All experiments are implemented in the PyTorch framework extending the QVNN library from \cite{parcollet2018quaternion}.\footnote{\url{https://github.com/Orkis-Research/Pytorch-Quaternion-Neural-Networks}} For replicability, we release our demo files on a separate repository online.\footnote{\url{https://github.com/Riccardo-Vecchi/Pytorch-Quaternion-Neural-Networks}} All hyper-parameters are fine-tuned independently for each network and dataset using the corresponding validation data. Importantly, this means that all regularization coefficients are optimized separately for every method.
 
 \subsection{Results for the quaternion-level sparsity}

\definecolor{LightCyan}{rgb}{0.88,1,1}
\begin{table*}[!ht]
	\small
	\ra{1.3}
	\centering
	\caption{Average results on MNIST and CIFAR-10 with several weight-level regularization strategies. With a light blue background, we highlight the (average) quaternion sparsity.} 
	\begin{tabular}{lllllll}
		\toprule
		Dataset & Measure & No Reg. & $\ell_2$ & $\ell_1$ & $R_Q$ & $R_{QL}$ \\ \midrule
		\multirow{3}{*}{MNIST} & Test accuracy [\%] & $98.95$ & $96.69$ & $93.46$ & $96.81$ & $96.29$ \\
		& Component sparsity [\%] & $1.71$ & $54.68$ & $34.08$ & $68.40$ & $75.08$ \\
		& \cellcolor{LightCyan!75}Quaternion sparsity [\%] & \cellcolor{LightCyan!75}$0.0$ & \cellcolor{LightCyan!75}$40.01$ & \cellcolor{LightCyan!75}$23.45$ & \cellcolor{LightCyan!75}$53.82$ & \cellcolor{LightCyan!75}$69.42$ \\ \midrule
		\multirow{3}{*}{CIFAR-10} & Test accuracy [\%] & $71.30$ & $72.58$ & $73.43$ & $72.03$ & $73.20$ \\
		& Component sparsity [\%] & $0.77$ & $8.58$ & $44.24$ & $59.29$ & $35.91$ \\
		& \cellcolor{LightCyan!75}Quaternion sparsity [\%] & \cellcolor{LightCyan!75}$0.0$ & \cellcolor{LightCyan!75}$4.77$ & \cellcolor{LightCyan!75}$42.61$ & \cellcolor{LightCyan!75}$58.73$ & \cellcolor{LightCyan!75}$31.89$ \\ \bottomrule
	\end{tabular}
	\vspace{0.5em}
	\label{tab:results}
\end{table*}

\begin{table*}[!ht]
	\small
	\ra{1.3}
	\centering
	\caption{Average training time (in seconds) for the different approaches.} 
	\begin{tabular}{llllll}
		\toprule
		Dataset & No Reg. & $\ell_2$ & $\ell_1$ & $R_Q$ & $R_{QL}$ \\ \midrule
		MNIST & $179.35$ & $183.17$ & $185.43$ & $184.22$ & $186.22$ \\ \midrule
		CIFAR-10 & $924.77$ & $1146.89$ & $1160.48$ & $1151.50$ & $1168.37$ \\ \bottomrule
	\end{tabular}
	\vspace{0.5em}
	\label{tab:training_time}
\end{table*}

We start by evaluating the quaternion-level regularization strategy described in Section \ref{subsec:l1_quaternion_regularization}, denoted as $R_Q$ in the experiments. We compare classical $\ell_2$ and $\ell_1$ regularizations, which are applied independently on every component. In addition, we evaluate a mixed regularization strategy combining our proposed $R_Q$ method with an additional $\ell_1$ regularization on the components, denoted as $R_{QL}$, which is similar to the \textit{sparse group sparse} technique in \cite{scardapane2017group}. For this case, we consider a single, shared regularization factor to be optimized to make comparisons fair.

\begin{figure*}[t]
\centering
\begin{subfigure}[c]{0.48\textwidth}
\includegraphics[width=0.98\linewidth]{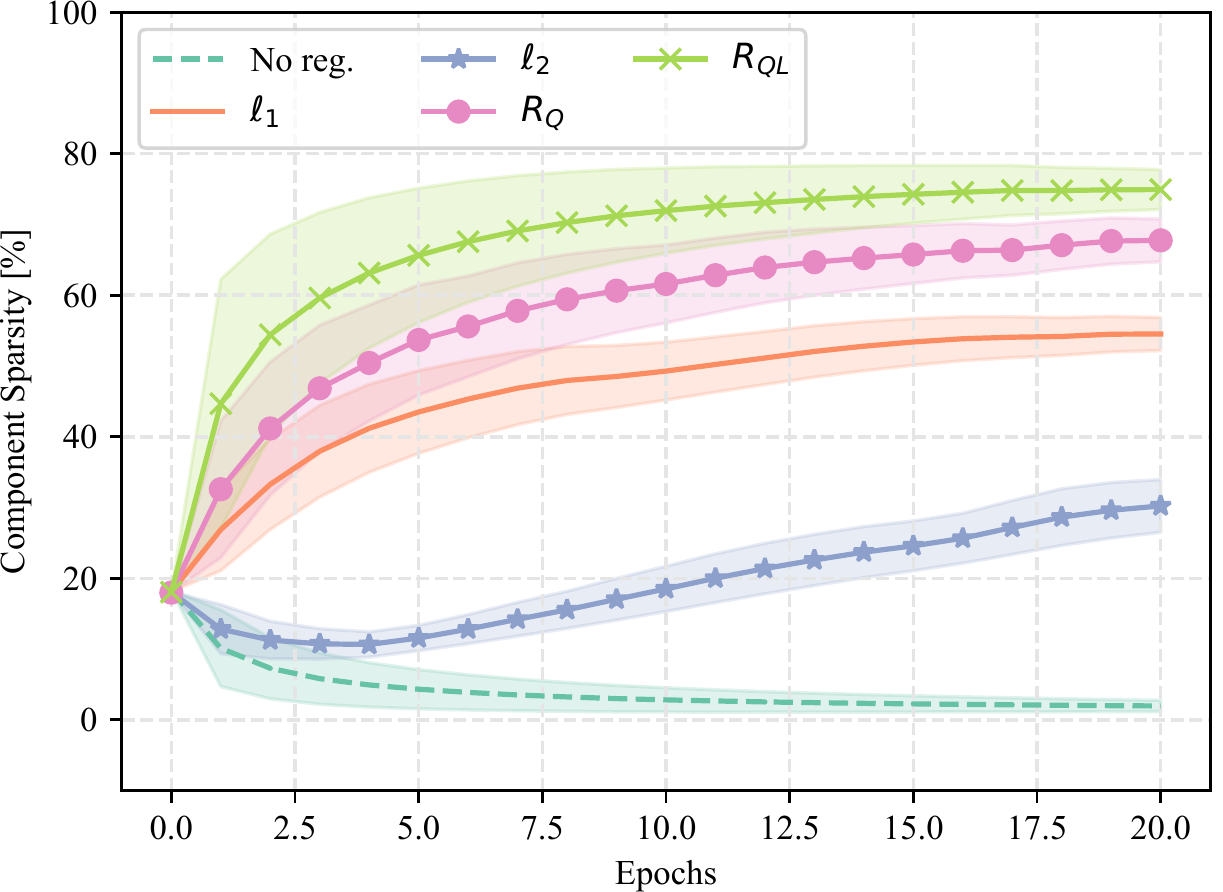}\label{fig:weight_sparsity_mnist}\caption{Component-level sparsity}
\end{subfigure}
\begin{subfigure}[c]{0.48\textwidth}
\includegraphics[width=0.98\linewidth]{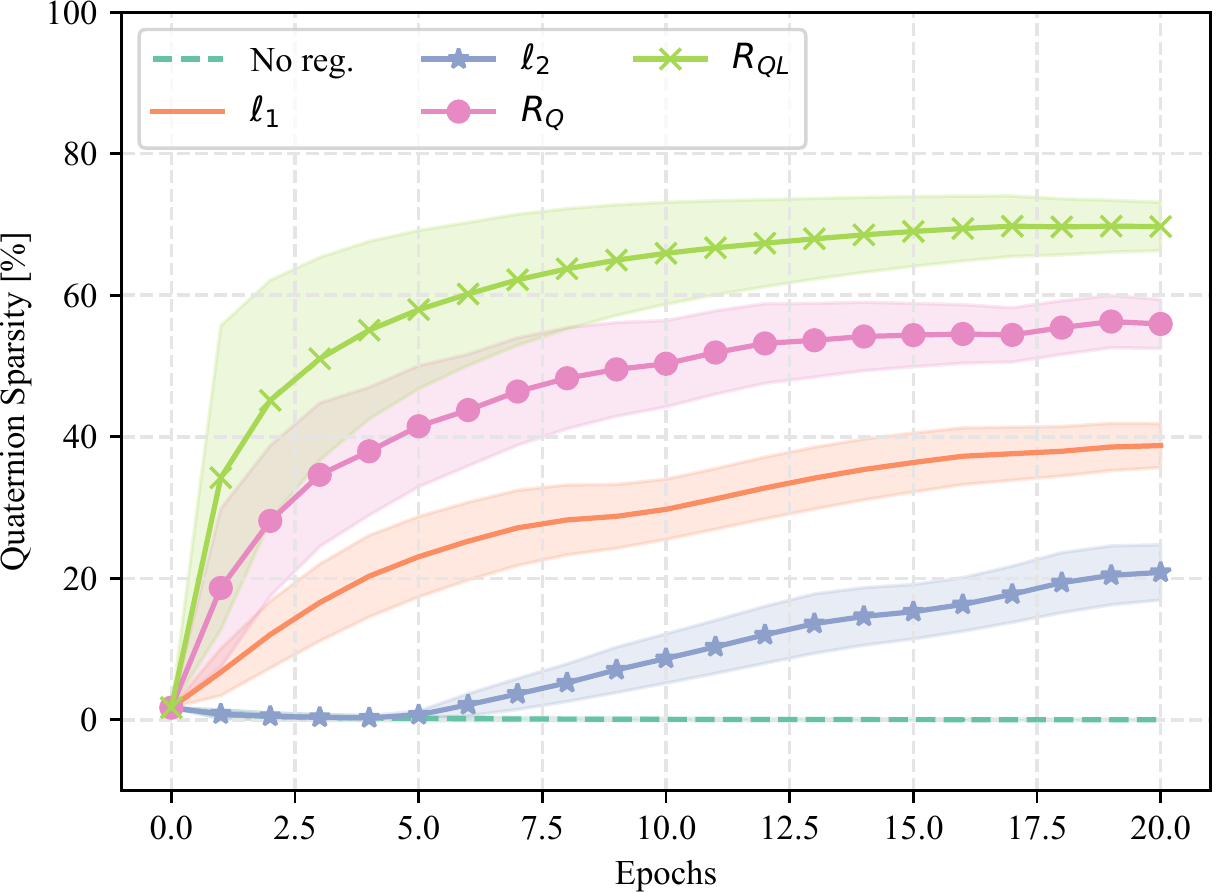}
\label{fig:quaternion_sparsity_mnist}
\caption{Quaternion-level sparsity}
\end{subfigure}
\caption{Evolution of the sparsity (both at the quaternion level and at the level of individual components) for the different strategies under consideration for the first $20$ epochs. See the text for a description of the acronyms. (a) Weight sparsity, (b) Quaternion-level sparsity.}
\label{fig:sparsity_mnist}
\end{figure*}

Results, averaged over $5$ different repetitions of the experiments, are presented in Tab. \ref{tab:results}. We see that applying a regularization has only a marginal effect on accuracy in the MNIST test accuracy, while it improves the accuracy in the more challenging CIFAR-10 case, possibly counter-acting any overfitting effect. In terms of sparsification effects, we show both the component sparsity (i.e., ratio of zero-valued quaternion components) and quaternion sparsity (i.e., ratio of quaternions where all components have been set to $0$). We can see that the proposed $R_Q$ strategy results in significantly sparser architectures in both cases, with corresponding gains when considering computational power and inference speed. The mixed strategy $R_{QL}$ performs very well on MNIST and poorer on CIFAR-10, possibly because we are using only a single shared regularization factor. For a clearer visualization, in Fig. \ref{fig:sparsity_mnist} we show the corresponding sparsity levels during training (for the first $20$ epochs of training).

In addition, in Tab. \ref{tab:training_time} we report the average training time on different experiments. As expected, adding a regularization term adds only a small overhead in terms of computational time, while we see no statistical difference between  different approaches, further validating the use of a targeted strategy.

\subsection{Results for the neuron-level sparsity}

\begin{figure*}[!ht]
\centering
\captionsetup[subfigure]{aboveskip=-3pt,belowskip=2pt}
\begin{subfigure}[t]{0.45\textwidth}
\includegraphics[width=\linewidth]{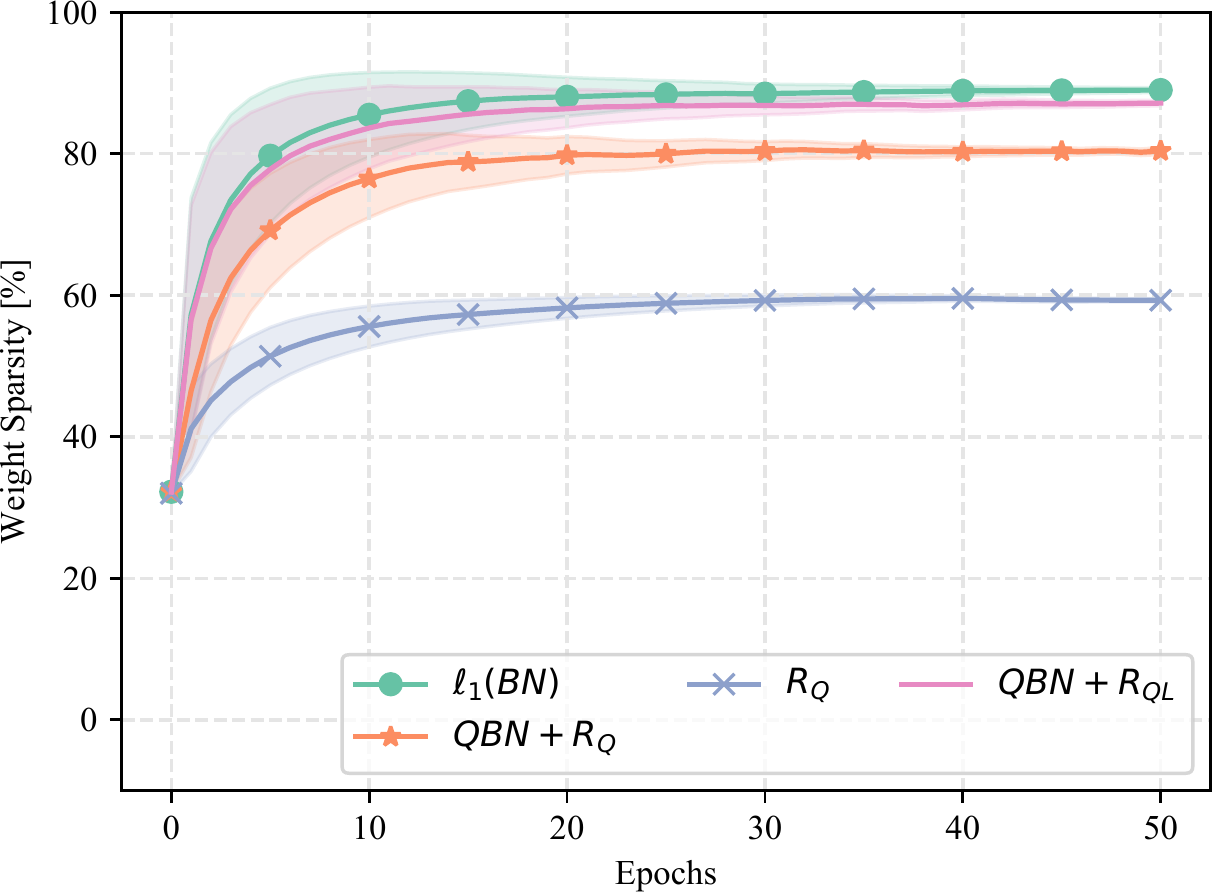}
\label{fig:CIFAR10_weight_sparsity}
\caption{Component-level sparsity}
\end{subfigure}
\begin{subfigure}[t]{0.45\textwidth}
\includegraphics[width=\linewidth]{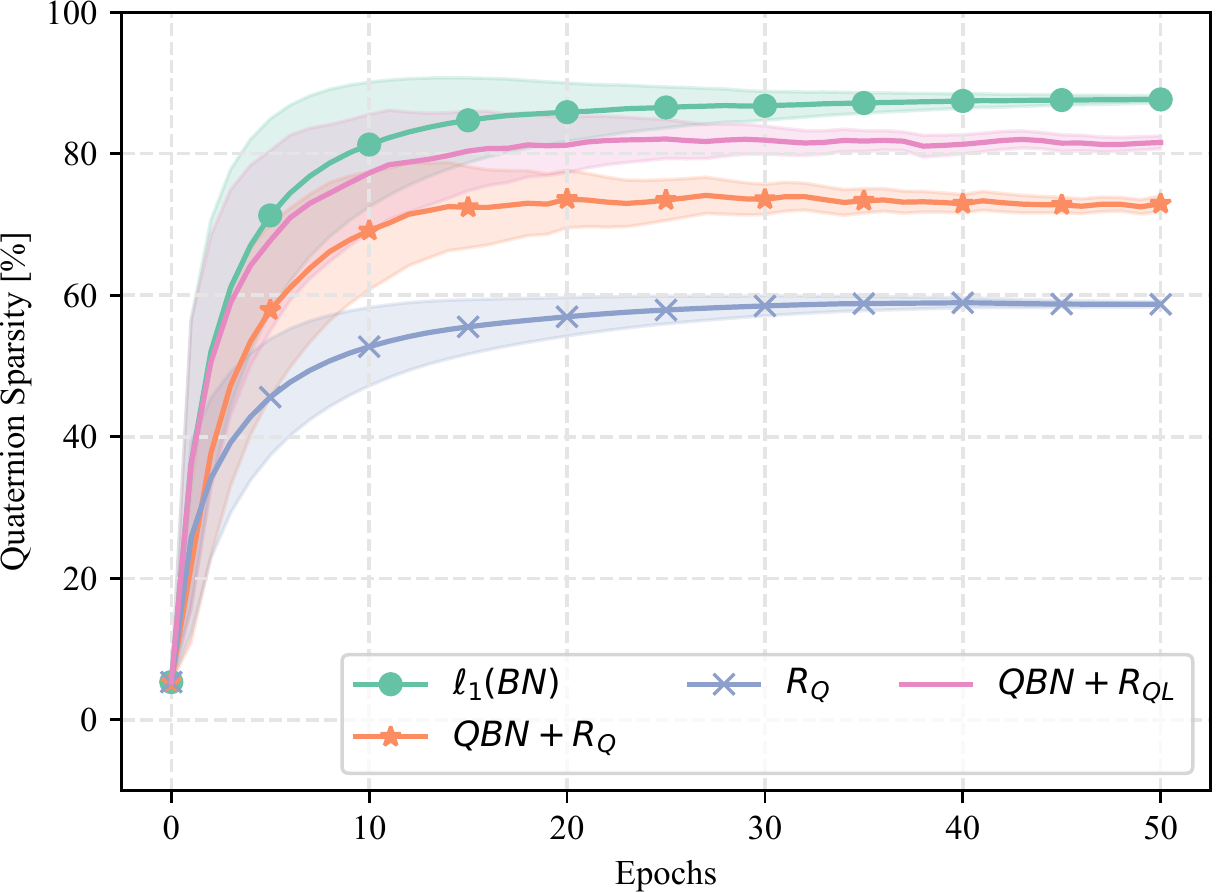}
\label{fig:CIFAR10_quaternion_sparsity}
\caption{Quaternion-level sparsity}
\end{subfigure}
\begin{subfigure}[t]{0.45\textwidth}
\includegraphics[width=\linewidth]{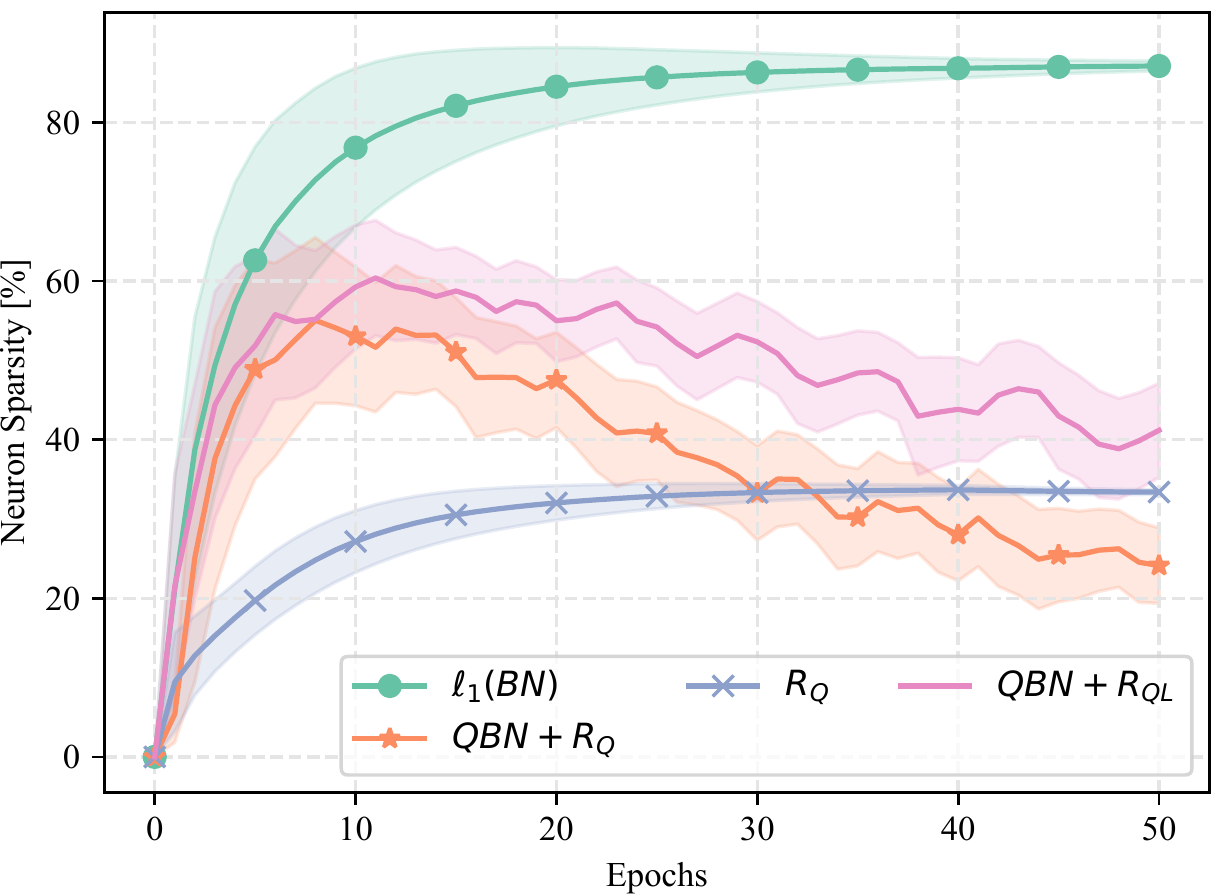}
\caption{Neuron-level sparsity}
\label{fig:CIFAR10_neuron_sparsity}
\end{subfigure}
\begin{subfigure}[t]{0.45\textwidth}
\includegraphics[width=\linewidth]{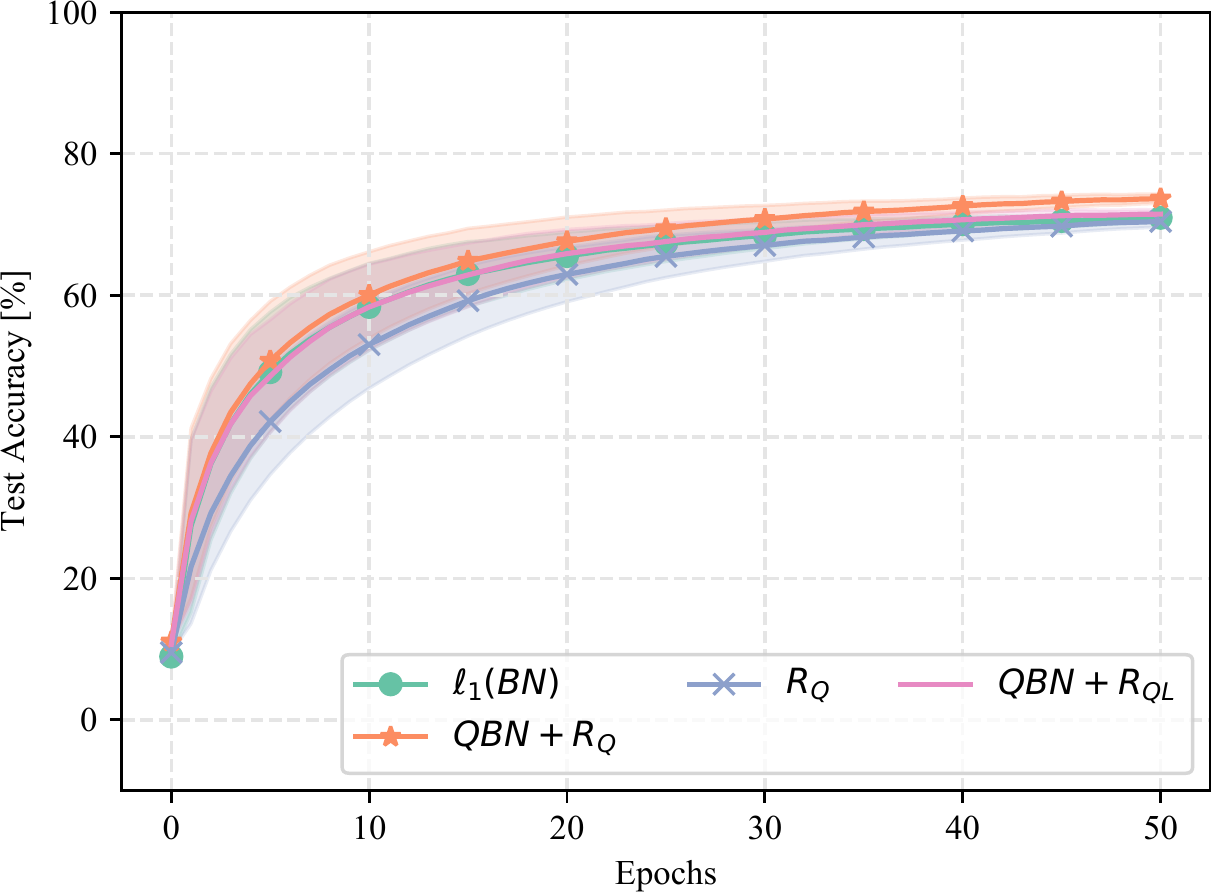}
\caption{Test accuracy}
\label{fig:CIFAR10_accuracy}
\end{subfigure}
\caption{Evolution of sparsity and accuracy for the CIFAR-10 dataset, zoomed on the first epochs. See the text for a description of the different acronyms. (a) Weight sparsity, (b) Quaternion-level sparsity, (c) Neuron sparsity, (d) Accuracy.}
\label{fig:sparsity_cifar10}
\end{figure*}

Next, we evaluate the inclusion of the neuron-level sparsity strategy described in Section \ref{subsec:sparse_regularization_with_quaternion_batch_normalization}. We consider the $R_Q$ strategy from the previous section, and compare a network where we add BN layers after every convolutive layer, penalizing the $\gamma$ coefficients with an $\ell_1$ strategy. For fairness, we also compare two additional baselines where we add the BN layers, but regularize only with the $R_Q$ or $R_{QL}$ strategies. For space constraints, we only consider the CIFAR-10 dataset, which was the most challenging in the previous section.

The averaged results are presented in Fig. \ref{fig:sparsity_cifar10}. We see that, when considering only quaternion-sparsity, the proposed neuron-level strategy (denoted as $L_1(BN)$) is marginally superior to the proposed $R_{QL}$ applied on the network having BN layers. However, when evaluating the level of structure in this sparsity, we see that the proposed neuron-level technique in Fig. \ref{fig:CIFAR10_neuron_sparsity} vastly outperforms all other strategies, leading to a network having less than $17\%$ of neurons than the original one, as long as having less than $15k$ remaining parameters. As a result, the final network has a memory footprint of only $1/5$ of the original one, with an inference time speedup of approximately 3x. From Fig. \ref{fig:CIFAR10_accuracy}, we also see that this is achieved with no loss in terms of test accuracy of the final networks and, similarly to Tab. \ref{tab:results}, with no significant increase in computational training time.

\section{Conclusions}
\label{sec:conclusions}

The field of quaternion neural networks explores the extensions of deep learning to handle quaternion-valued data processing. This has shown to be especially promising in the image domain and similar fields with highly structured data that lend itself to representation in a hyper-complex domain. While several models and training algorithms have been already extended to this new challenging domain, less attention has been provided to the tasks of regularizing and compressing the networks, which is essential in time-critical and embedded applications. 

In this paper, we proposed two regularization techniques that are specific to quaternion-valued networks. In the first case, we apply some results from quaternion compressive sensing, regularizing each quaternion weight with an $\ell_1$-level norm. In the second case, we consider the problem of removing entire neurons from the network, by regularizing appropriately inserted batch normalization layers. Our experimental results on two image classification benchmarks show that these two techniques vastly outperform standard regularization methods when they are applied to quaternion networks, allowing to obtain networks that are extremely smaller (and cheaper to implement) with no loss in testing accuracy at inference time and in computational cost at training time.

For future work, we plan on extending these results to other applications of quaternion-based deep networks (e.g., text processing), as long as evaluating the benefits of sparsifying the networks on custom hardware.

\bibliographystyle{ieeetr}
\bibliography{biblio}

% that's all folks
\end{document}